\PassOptionsToPackage{sort}{natbib}

\PassOptionsToPackage{linesnumbered,ruled,vlined}{algorithm2e}

\PassOptionsToPackage{hidelinks}{hyperref}

\documentclass[pmlr,twocolumn,10pt]{jmlr} 

\usepackage{xurl}  
\usepackage{cuted}      
\usepackage{caption} 

\usepackage{booktabs,array,makecell,graphicx}
\newcolumntype{L}[1]{>{\raggedright\arraybackslash}m{#1}} 
\newcolumntype{C}[1]{>{\centering\arraybackslash}m{#1}}   
\newcommand{\NA}{\textemdash}
\newcommand{\tablescale}{0.85}

\DontPrintSemicolon
\SetKwInput{KwInput}{Input}
\SetKwInput{KwOutput}{Output}
\SetKwComment{Comment}{$\triangleright$ }{}
\makeatletter
\@ifpackageloaded{algorithm2e}{}{\usepackage{algorithm2e}}
\makeatother

\RestyleAlgo{ruled}           
\SetAlgoVlined                
\DontPrintSemicolon

\newcolumntype{P}[1]{>{\raggedright\arraybackslash}p{#1}}

\usepackage{booktabs}
\usepackage{siunitx}

\usepackage[switch]{lineno}

\usepackage{enumitem}
\usepackage{caption}

\theorembodyfont{\upshape}
\theoremheaderfont{\scshape}
\theorempostheader{:}
\theoremsep{\newline}

\jmlrvolume{297}
\jmlryear{2025}
\jmlrworkshop{Machine Learning for Health (ML4H) 2025} 

 \title[Short Title]{ML4H 2025 Template: Proceedings Track}

\title{One VLM, Two Roles: Stage-Wise Routing and Specialty-Level Deployment for Clinical Workflows}

\author{%
\Name{Shayan Vassef}\Email{shayan@nimblemind.ai}\\
\addr Nimblemind, USA
\AND
\Name{Soorya Ram Shimgekar}\Email{soorya@nimblemind.ai}\\
\addr Nimblemind, USA
\AND
\Name{Abhay Goyal}\Email{abhay@nimblemind.ai}\\
\addr Nimblemind, USA
\AND
\Name{Christian Poellabauer}\Email{cpoellab@fiu.edu}\\
\addr Florida International University, FL, USA
\AND
\Name{Koustuv Saha}\Email{ksaha2@illinois.edu}\\
\addr University of Illinois - Urbana Champaign, IL, USA
\AND
\Name{Pi Zonooz}\Email{pi@nimblemind.ai}\\
\addr Nimblemind, USA
\AND
\Name{Navin Kumar} \Email{navin@nimblemind.ai}\\
\addr Nimblemind, USA
}

\begin{document}

\maketitle
\begin{abstract}
Clinical ML workflows are often fragmented and inefficient: triage, task selection, and model deployment are handled by a patchwork of task-specific networks. These pipelines are rarely aligned with data-science practice, reducing efficiency and increasing operational cost. They also lack data-driven model identification (from imaging/tabular inputs) and standardized delivery of model outputs. We present a framework that employs a \emph{single} vision--language model (VLM) in two complementary, modular roles.
\textbf{First (Solution~1):} the VLM acts as an aware model-card matcher that routes an incoming image to the appropriate specialist model via a three-stage workflow (modality $\rightarrow$ primary abnormality $\rightarrow$ model-card ID). Reliability is improved by (i) stage-wise prompts enabling early termination via \texttt{None}/\texttt{Other} and (ii) a calibrated top-2 answer selector with a stage-wise cutoff. This raises routing accuracy by \textbf{+9} and \textbf{+11} percentage points on the training and held-out splits, respectively, compared with a baseline router, and improves held-out calibration (lower ECE).
\textbf{Second (Solution~2):} we fine-tune the same VLM on specialty-specific datasets so that one model per specialty covers multiple downstream tasks, simplifying deployment while maintaining performance. Across gastroenterology, hematology, ophthalmology, pathology, and radiology, this single-model deployment \textbf{matches or approaches} specialized baselines.

Together, these solutions reduce data-science effort through more accurate selection, simplify monitoring and maintenance by consolidating task-specific models, and increase transparency via per-stage justifications and calibrated thresholds. Each solution stands alone, and in combination they offer a practical, modular path from triage to deployment.
\end{abstract}

\begin{figure*}[t]
  \centering
  \includegraphics[width=1\textwidth]{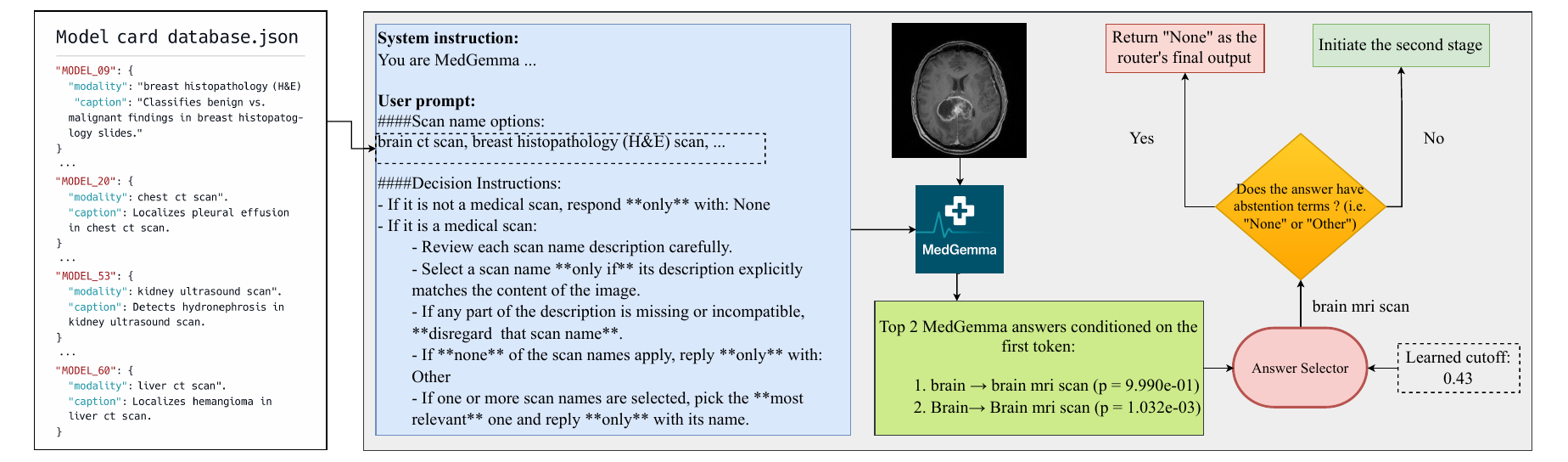}
  \caption{The first stage in the router workflow (modality detection).}
  \label{fig:stage1}
\end{figure*}

\begin{figure*}[t]
  \centering
  \includegraphics[width=0.75\textwidth]{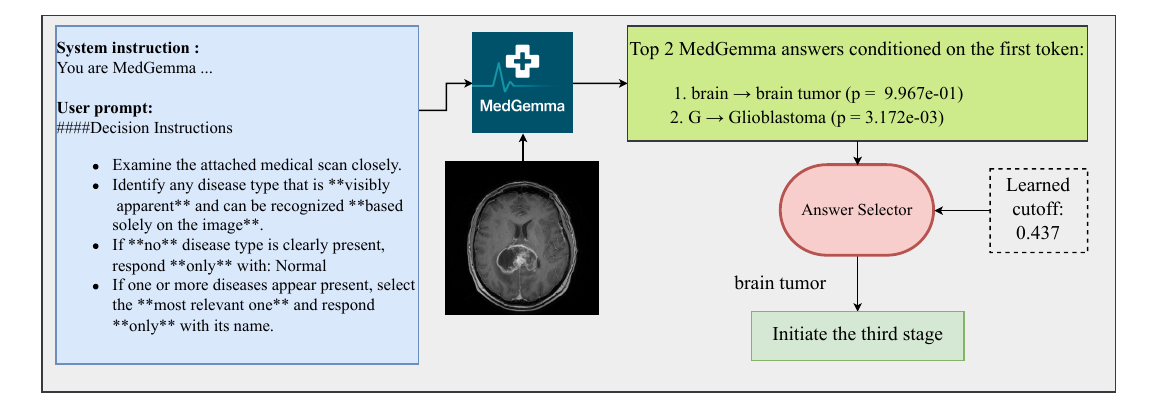}
  \caption{The second stage in the router workflow (disease identification).}
  \label{fig:stage2}
\end{figure*}

\begin{figure*}[t]
  \centering
  \includegraphics[width=0.75\textwidth]{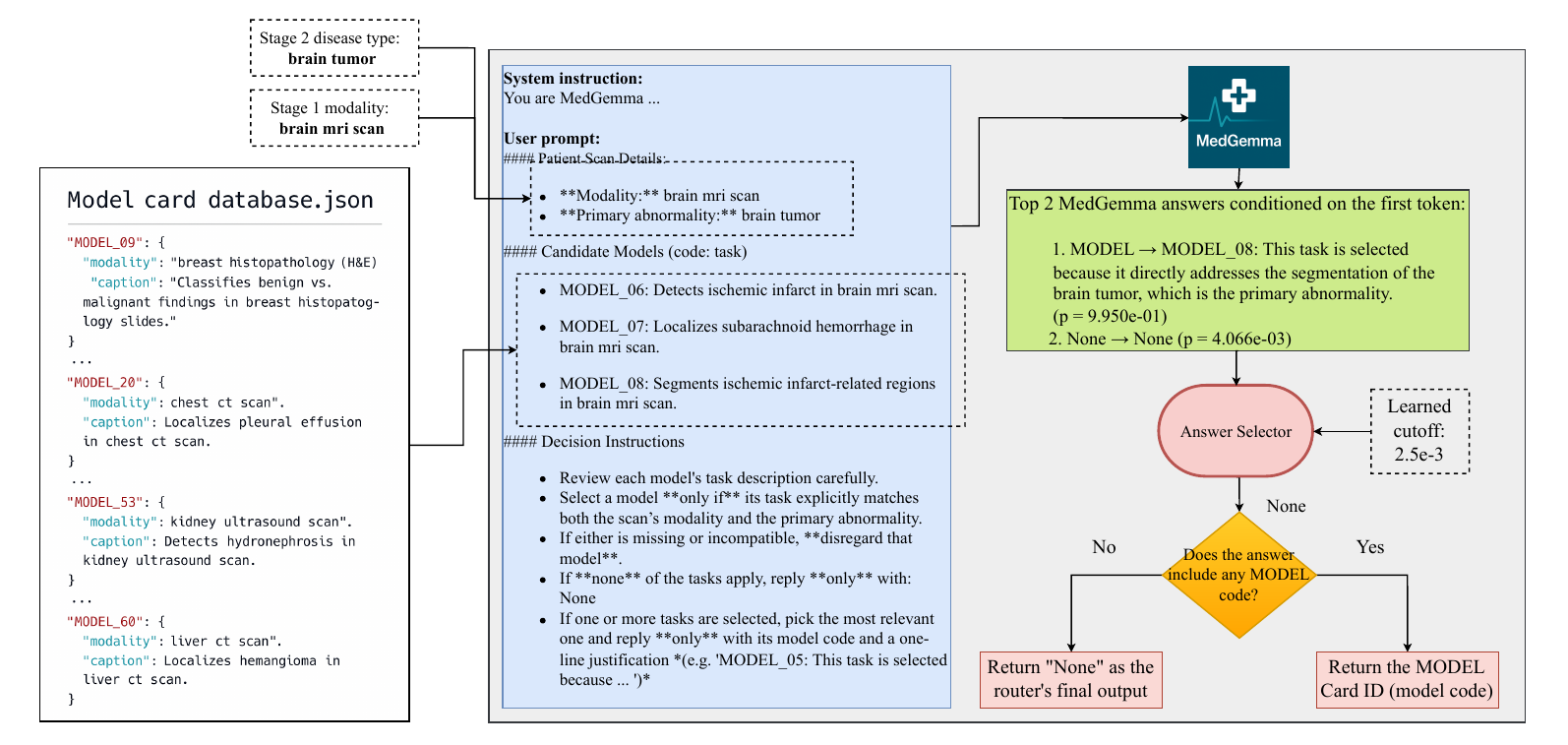}
  \caption{The third (final) stage in the router workflow (model-card matching).}
  \label{fig:stage3}
\end{figure*}


\paragraph*{Data and Code Availability}
Data and code are available from the authors at reasonable request. 

\paragraph*{Institutional Review Board (IRB)}
All data is publicly available and de-identified; the study is IRB-exempt. 

\section{Introduction}
\label{intro}
Machine learning (ML) models in healthcare face significant operational friction when moving from research prototypes to clinical practice. An estimated 90\% of medical AI models never reach clinical deployment~\citet{poddar2024translational}, and production-ready models can take weeks to months to integrate, with over half of organizations reporting 8–90 days to deploy a single model~\citet{mage2025mlmodeltime}. These delays stem from several bottlenecks, and we provide two key examples here.
First, to select an appropriate model design (architecture, training objective, etc.) for a given dataset, the data scientist must identify the attributes that guide model design. In tabular data these attributes are explicit; in clinical images they are latent in the pixels—such as \emph{modality} (e.g., CT) and \emph{clinical indication} (e.g., acute intracranial hemorrhage)—and must first be extracted, standardized, and quality-checked before they can inform design choices. As a result, streamlined model selection for clinical imaging is often overlooked and warrants dedicated attention~\citet{willemink2020_preparing}.
Second, health systems tend to integrate, validate, monitor, and recalibrate AI solutions one by one, so developing and maintaining task-specific models multiplies the cost burden and slows adoption~\citet{brady2024developing}.

We address these operational issues with two linked—but independently evaluated—solutions: the first assists data scientists in selecting the appropriate model, and the second shortens time-to-deployment by unifying task-specific AI models into broader, specialty-level models that support multiple tasks. While Solution~1 could inform Solution~2 in future systems, \emph{our implementation treats them separately}; Solution~2 does not consume outputs from Solution~1.

\paragraph{\textbf{Solution~1: Model-card matching}}
Early medical vision–language efforts aligned images and text to enable cross-modal understanding and transfer. Recently, MedGemma~\citet{sellergren2025medgemma} and UMIT~\citet{yu2025umit} demonstrated that medical VLMs can approach specialized systems while maintaining generality, motivating workflows that both interpret the input and decide what to do next. Accordingly, we adopt a medical VLM as an aware, auditable selector that reads an input image and chooses an appropriate model card via a three-stage pipeline (\textsc{modality} $\rightarrow$ \textsc{primary abnormality} $\rightarrow$ \textsc{model-card ID}). At each stage, the VLM proposes top candidates, and an answer selector applies a calibrated threshold that either selects the most likely candidate (our baseline) or, if warranted, the second most likely candidate. Decisions at each stage (VLM output, any abstention, and routing to the next stage) are visible to users and logged in a consistent format, yielding transparent selection tied to model-card semantics~\citet{mitchell2019modelcards} and safe selective prediction~\citet{geifman2019selectivenet}. This process accelerates development: we posit that choosing the appropriate model for a given clinical image hinges on two key pieces of information—imaging modality and disease type—details that are not always apparent to data scientists.

\paragraph{\textbf{Solution~2: Specialty-specific deployment}}
Prior work shows that domain-pretrained medical encoders, when fine-tuned on target tasks, can match strong baselines across multiple datasets~\citet{zhang2022_convirt,mei2022_radimagenet}. In healthcare, specialty-level foundation models have also demonstrated broad coverage within a specialty—outperforming narrow, dataset-specific models across many tasks (e.g., pathology’s Prov-GigaPath achieves SOTA on 25/26 tasks across 31 tissue types)~\citet{xu2024whole}. Operationally, fewer, broader specialty models reduce the number of artifacts a health system must validate, secure, and monitor, lowering deployment burden compared with one-model-per-dataset estates~\citet{brady2024developing}. In line with past work, we fine-tune the \emph{same} calibrated VLM at the \emph{specialty} level—gastroenterology, hematology, ophthalmology, pathology, and radiology—rather than per dataset. Our goal is to match task-specific model performance while keeping everything on a single, easy-to-maintain deployment stack. To adapt the VLM to multiple AI tasks, one must include task-specific prompts in the VLM’s training data (e.g., a task-specific user prompt for colonoscopy lesion classification and a task-specific user prompt for polyp bounding-box detection) to build a specialty model that supports both classification and bounding-box (BBox) prediction. Importantly, in our experiments, \emph{Solution~2 is trained and evaluated independently of Solution~1}.

Taken together, these solutions connect selection and deployment in a calibrated workflow—a VLM that can \emph{route and execute}. Compared with recent multi-agent pipelines~\citep{shimgekar2025agentic, 202508.1713}, this minimalist design may reduce data-science workload and human error. As an example, AI-driven triage has been shown to cut chest-radiograph reporting delays from 11.2 to 2.7 days in simulation while maintaining diagnostic performance~\citet{annarumma2019triage}. Similarly, our approach may shorten the path from triage to clinical use and increase throughput without sacrificing performance.

\section{Related Work}
\paragraph{\textbf{Automated matching in healthcare}}
Selecting the appropriate model based on a dataset has been explored at multiple layers of care delivery, such as patient-facing systems, specialty care, etc. Patient-facing systems (e.g., symptom checkers like Buoy Health ~\citet{Buoy}) narrow likely conditions and suggest next steps. Platforms such as Garner Health ~\citet{Garner} match patients to specialists using outcomes data. Similarly, LLM-driven controllers (HuggingGPT ~\citet{shen2023hugginggpt}) show how a coordinator can parse requests, consult model descriptions, and delegate to expert models. This pattern maps naturally to healthcare \emph{model cards} as standardized summaries of intended use and performance~\citet{mitchell2019modelcards}. Clinical imaging ``orchestration” seeks to run the appropriate algorithm on the selected scan \cite{aidoc_orchestration}. However, prior work does not provide calibrated, stage-wise routing that is explicitly grounded in model-card semantics, with visible abstention and early termination and per-stage decision logs for audit. In the first solution, we address this gap with a three-stage router that makes every decision—VLM output, abstention, and route—user-visible and logged, cutting routing errors while preserving safety via careful prompt design and stage-specific selection logic.

\vspace{-2\baselineskip}
\paragraph{\textbf{Adaptation across tasks}}
A parallel line of work reduces packaging and integration friction for \emph{multiple} models: MONAI ~\citet{monai_deploy} deploys standardized applications, packaging and interfaces, and self-configuring pipelines like nnU-Net ~\citet{isensee2021nnunet} that adapt architectures to new datasets with minimal manual tuning. In practice, however, maintaining one network per task is inefficient: each model has to be validated, integrated, monitored, and updated separately at the site level, duplicating effort and slowing adoption~\citet{brady2024developing, boverhof2024radiology}.
Multi-agent or model-zoo deployments likewise increase the number of components to secure, monitor, and recalibrate over time, compounding technical debt~\citet{sculley2015mldebt}.
Unlike UMIT \cite{yu2025umit}—which initializes Qwen2-VL \citet{wang2024qwen2} and follows a two-stage pipeline (full-parameter feature alignment followed by instruction fine-tuning) to reach competitive results across many datasets and modalities—there is no evidence that a single calibrated VLM can be \emph{reused} across specialties within one unified deployment stack where onboarding a new task changes only the \emph{user prompt}. Our second solution targets this gap: we fine-tune one medical VLM across datasets within a specialty so a single component serves multiple imaging tasks, keeping interfaces and calibration fixed while matching or approaching benchmark performance.




\section{Solution 1: Model--Card Matching with Early Termination}
\label{sol1}
To integrate the medical VLM as an automated model selector, we employ a three-stage prompting pipeline. At each stage, the model receives a stage-specific instruction and context and returns a textual answer. The stages progressively narrow the information needed to select the appropriate model card. After each response, an \textbf{answer-selection} step ranks candidate answers by likelihood, retains the top two, and chooses between them; for Stage~1 only, it may also terminate early. Figures~\ref{fig:stage1}, \ref{fig:stage2}, and \ref{fig:stage3} illustrate the overall workflow in order; Algorithm~\ref{alg:answer-selector} provides pseudocode for the answer-selection strategy; and Algorithm~\ref{alg:cutoff-sweep} explains how we derive stage-wise cutoffs that parameterize Algorithm~\ref{alg:answer-selector}. Below, we first outline the prerequisites and then describe the selection process for each stage.

\paragraph{\textbf{Prerequisite~1: Backbone VLM choice}}
In Section~\ref{intro} (Solution~1) we described a generic medical VLM; here, we \emph{adopt} \textsc{MedGemma}. MedGemma was pretrained on a multimodal medical mixture spanning radiology (CXR/CT/MRI), ophthalmology, dermatology, and histopathology\footnote{Example datasets: MIMIC-CXR 231{,}483 images; EyePACS 199{,}258; internal dermatology 51{,}049; PAD-UFES-20 2{,}047; internal histopathology 32.55M patches; CT-US1 59{,}979 slices; MRI-US1 47{,}622~\citet{sellergren2025medgemma}.} and reports competitive benchmarks relative to prior work~\citet{sellergren2025medgemma}.

\paragraph{\textbf{Prerequisite~2: Model-card repository}}
To match a clinical image to the correct model, we maintain a diverse repository of model cards spanning multiple modalities (e.g., CT, MRI) and AI tasks (e.g., classification, grading). This repository currently includes \textbf{84} model cards over \textbf{26} unique modalities: brain CT scan, brain MRI scan, breast histopathology (H\&E) scan, breast mammography scan, breast ultrasound scan, cervix cytology scan, cervix histology scan, cervix histopathology (H\&E) scan, chest CT scan, colonoscopy scan, colon histopathology (H\&E) scan, esophageal gastroscopy scan, fundus photography scan, gallbladder ultrasound scan, kidney CT scan, kidney histopathology scan, kidney MRI scan, kidney ultrasound scan, liver CT scan, liver ultrasound scan, lung histopathology (H\&E) scan, prostate histopathology (H\&E) scan, skin dermatology scan, skin dermoscopy scan, skin histopathology (H\&E) scan, and a skin multimodal panel (dermoscopy+RCM+H\&E). Rather than storing task-specific models, each candidate is indexed by a string key—the \emph{model-card ID} (a.k.a.\ model code; e.g., \texttt{MODEL\_04}, which might refer to an ophthalmology model)—and by two minimal descriptors that make it discoverable by the router: (i) a short \emph{task caption} describing what the model does (e.g., “segments tumor regions in histopathology images”), and (ii) the \emph{modality}, i.e., the scan type on which the model was trained (e.g., “breast histopathology scan”).

\paragraph{\textbf{Stage~1: Modality identification (Figure~\ref{fig:stage1}).}}
The goal of the first stage is to determine the imaging modality or scan type of the input image. We prompt MedGemma with a system message that casts it as a medical-imaging specialist and supply a list of admissible modalities drawn from the model-card repository. Concretely, we iterate over all model cards, extract each card’s \texttt{modality} field, deduplicate the values to form a unique set, and inject that set into the prompt.
The prompt instructs: (1) if the image is \emph{not} a medical scan, respond with \texttt{None}; (2) if it \emph{is} a medical scan, choose the single most appropriate scan name from the list that matches the image content; and (3) if none of the listed modalities perfectly fits, respond with \texttt{Other}. This prompting strategy explicitly conditions the model to use the tokens \texttt{None} or \texttt{Other} when a confident match is absent. For example, given an endoscopic image of the colon, the correct output might be “colon colonoscopy scan”; for a non-medical photograph (e.g., a picture of a cat), the model should output \texttt{None}. Early termination occurs only at this stage: if the selected token is \texttt{None} or \texttt{Other}, the pipeline halts.

\paragraph{\textbf{Stage~2: Primary finding detection (Figure~\ref{fig:stage2}).}}
In the second stage, the model examines the image (unaware of the router’s result for Stage~1) and attempts to identify the primary abnormality or finding, if any. We prompt MedGemma with an instruction such as: “Identify the single most likely disease or finding apparent in this scan. If no disease is present, respond only with: ‘Normal.’” This stage is essentially an image-based diagnosis step for a high-level finding. For example, given a colonoscopy image, the model might answer “Polyp” if one is seen; given a chest X-ray, it might answer “Normal” if nothing abnormal stands out, or a broad condition like “Pneumonia” if detected. The prompting ensures that the special token “Normal” is used to indicate the absence of any visible pathology.

\paragraph{\textbf{Stage~3: Model-card selection (Figure~\ref{fig:stage3}).}}
In the final stage, the model selects an appropriate card from the repository. We iterate over all cards, filter to those whose \texttt{modality} matches the Stage~1 choice, and pass the resulting list of \texttt{(model\_card\_id, task\_caption)} pairs to the prompt.
For example, if Stage~1 predicts \texttt{breast histopathology scan}, the candidate set might be:
[\texttt{MODEL\_01}: \texttt{Classifies benign vs.\ malignant findings in breast histopathology slides.},
 \texttt{MODEL\_04}: \texttt{\dots}].

To provide sufficient context for disambiguation, we include the Stage~1 modality and the Stage~2 primary abnormality as structured fields in the prompt (e.g., \emph{Patient scan details}: breast histopathology scan; \emph{Primary abnormality}: carcinoma). To make this a pure selection task, we mask the input image so that MedGemma acts as a language model solving a multiple-choice problem over the candidate list. The prompt instructs the model to: (i) review each candidate’s task caption; (ii) select a model only if its task explicitly matches \emph{both} the identified modality and abnormality; (iii) reply \texttt{None} if no option fits; and (iv) when a match exists, return the model’s code (card ID) plus a one-line justification of the fit.

\paragraph{\textbf{Cutoff Tuning and Answer Selection.}}
Even with careful prompting (e.g., \texttt{return the best-matched}, restrictive cues like \texttt{only}, and explicit abstentions such as \texttt{None}), the router can still pick suboptimal first tokens or miss safe abstentions. Empirically, we observe that the \emph{second-most likely} first token can sometimes yield a more faithful full answer. We therefore arbitrate between the top two candidates using a simple, stage-specific \emph{cutoff} on the runner-up probability.

\paragraph{\textbf{Selector logic.}}
At each stage, the model forms a distribution over the first token. We take the top two tokens and decode full answers conditioned on each, producing $\{(a_i,p_i)\}_{i=1}^{2}$ with $p_1\!\ge p_2$. This pair is passed to a transparent selector that prefers the runner-up only when its probability is sufficiently large. The rule returns both the chosen answer $\hat a$ and the index $y\in\{\textsc{first},\textsc{second}\}$. When applying Algorithm~\ref{alg:answer-selector} \textbf{in Stage~1 only}, if the selected token is literally \texttt{None} or \texttt{Other}, we halt and return \texttt{None} as the router’s final output (early termination to avoid extra compute).

\RestyleAlgo{ruled}
\SetAlgoNlRelativeSize{-1}
\SetKwInput{KwGiven}{Given}
\SetKwInput{KwOutput}{Output}

\begin{algorithm}
\floatconts
  {alg:selector}
  {\caption{Answer selector (per stage)} \label{alg:answer-selector}}
{%
\footnotesize
\setlist[enumerate]{leftmargin=1.6em,itemsep=2pt,topsep=3pt,parsep=0pt}
\setlist[enumerate,2]{leftmargin=1.4em,itemsep=0pt,topsep=0pt}

\begin{tabular}{@{}l l@{}}
\textbf{Given}  & $\mathcal{C}=\{(a_i,p_i)\}_{i=1}^{|V|}$ with $p_1\!\ge\!p_2$, cutoff $\tau$.\\
\textbf{Output} & $\hat{a}$ (selected answer), $y \in \{\textsc{first}, \textsc{second}\}$.
\end{tabular}

\begin{enumerate}
  \item Set $\hat a\!\gets a_1$ and $y\!\gets\textsc{first}$.
  \item If $p_2\!\ge\!\tau$ and $a_2\!\neq\!a_1$ then
    \begin{enumerate}
      \item[] set $\hat a\!\gets a_2$ and $y\!\gets\textsc{second}$ \emph{(prefer runner-up when $p_2$ is large)}.
    \end{enumerate}
\end{enumerate}
\vspace{-2pt}
}%
\end{algorithm}

\providecommand{\cutoff}{cutoff}
\providecommand{\Cutoff}{Cutoff}

\paragraph{\textbf{Learning the cutoff in a supervised fashion.}}
We gathered 102 diverse (image, per-stage top-2 predictions, label \emph{(First/Second)}) pairs over the training set and 18 pairs over the held-out set. All samples were collected from Wikimedia Commons and share a Creative Commons Attribution–ShareAlike (CC BY-SA) license. For each stage $k\!\in\!\{1,2,3\}$, the training split provides the correct index ($\textsc{first}$ vs.\ $\textsc{second}$).
We learn a single \emph{cutoff} $\tau_k$ via a one-pass sweep over the sorted runner-up probabilities $p_2$, which returns \emph{all} optimal intervals and a canonical midpoint for deployment. Applying Algorithm~\ref{alg:cutoff-sweep} with the $p_2$-based selector (Alg.~\ref{alg:answer-selector}) yields the optimal intervals and representative \cutoff{}s in Table~\ref{tab:cutoff-intervals}. Intervals are open on the right; wide plateaus indicate robustness.

\begin{algorithm}
\floatconts
  {alg:sweep}
  {\caption{Cutoff sweep (training split, stage $k$)} \label{alg:cutoff-sweep}}
{%
\footnotesize
\setlist[enumerate]{leftmargin=1.6em,itemsep=2pt,topsep=3pt,parsep=0pt}
\setlist[enumerate,2]{leftmargin=1.4em,itemsep=0pt,topsep=0pt}

\begin{tabular}{@{}l l@{}}
\textbf{Given}  & Training rows for stage $k$ with runner-up \\ &probabilities $p_2$ and labels $y\!\in\!\{\textsc{first},\textsc{second}\}$.\\
\textbf{Output} & All optimal \cutoff{} intervals for $\tau_k$; canonical $\tau_k^\star$ \\& (midpoint of the widest interval).
\end{tabular}

\begin{enumerate}
  \item \textbf{Partition events:} $A\!\leftarrow\!\{p_2:\,y=\textsc{second}\}$,\quad $B\!\leftarrow\!\{p_2:\,y=\textsc{first}\}$.
  \item \textbf{Sort once:} merge $A\cup B$ ascending, consuming equal values together (event list).
  \item \textbf{Initialize score:} place $\tau$ just left of $0$; $S_0\!\gets\!|A|$.
  \item \textbf{One-pass sweep (left$\rightarrow$right):}
    \begin{enumerate}
      \item Crossing an $A$-value: a previously correct \emph{flip-to-second} becomes incorrect $\Rightarrow S\!\gets\!S-1$.
      \item Crossing a $B$-value: a previously incorrect \emph{stay-with-first} becomes correct $\Rightarrow S\!\gets\!S+1$.
    \end{enumerate}
  \item \textbf{Track maxima:} record $S_{\max}$ and every open interval between successive events where $S=S_{\max}$.
  \item \textbf{Return:} these intervals are optimal; set $\tau_k^\star$ to the midpoint of the \emph{widest} optimal interval.
\end{enumerate}
\vspace{-2pt}
}%
\end{algorithm}


\vspace{-1.5\baselineskip}
\begin{table}[!ht]
  \centering
  \scriptsize
  \setlength{\tabcolsep}{3pt}
  \caption{Optimal cutoff intervals and representative $\tau_k^\star$
    learned on the training split.}
  \label{tab:cutoff-intervals}
  \begin{tabular}{@{}l P{0.45\linewidth} c@{}}
    \toprule
    \textbf{Stage} & \textbf{Optimal interval(s)} & \textbf{$\tau_k^\star$} \\
    \midrule
    Stage 1 — Modality &
      \texttt{[0.417100, 0.434750)} &
      \textbf{0.425925} \\
    Stage 2 — Disease  &
      \texttt{[0.436800, 0.437400)} &
      \textbf{0.437100} \\
    Stage 3 — Final    &
      \begin{tabular}[t]{@{}l@{}}
        \texttt{[0.001100, 0.001334)};\\
        \texttt{[0.001985, 0.003030)}
      \end{tabular}
      & \textbf{0.002508} \\
    \bottomrule
  \end{tabular}
\end{table}

\vspace{-1.2\baselineskip}
\paragraph{\textbf{Per-stage accuracy (training and held-out).}}
We compare a simple baseline (always choose \textsc{first}) with the \cutoff{} router (Alg.~\ref{alg:answer-selector} using $\tau_k^\star$ from Alg.~\ref{alg:cutoff-sweep}). The $p_2$-based selector is transparent and stage-local. Cutoffs learned by a single sweep improve final-stage correctness on both splits (approximately +10 pp on training split; approximately +6 pp on held-out split), with modest gains at earlier stages. The inference-time Stage~1 abstention guard reduces unnecessary computation without affecting cutoff learning or the evaluations above.


\vspace{-0.2\baselineskip} 
\begin{table}[!ht]
  \centering
  \scriptsize
  \setlength{\tabcolsep}{2pt}
  \caption{Per-stage accuracy with and without \cutoff{}.}
  \label{tab:per-stage-true}
  \begin{tabular}{@{}llcc@{}}
    \toprule
    \textbf{Split} & \textbf{Stage / Task} & \textbf{Baseline acc} &
    \textbf{\Cutoff{} acc (ours)} \\
    \midrule
    Training & Stage 1 — Modality & 0.9216 & \textbf{0.9412} \\
             & Stage 2 — Disease  & 0.7647 & \textbf{0.7745} \\
             & Stage 3 — Final    & 0.4608 & \textbf{0.5588} \\
    \midrule
    Held-out & Stage 1 — Modality & 0.8889 & \textbf{0.8889} \\
             & Stage 2 — Disease  & 0.6667 & \textbf{0.6667} \\
             & Stage 3 — Final    & 0.5556 & \textbf{0.6111} \\
    \bottomrule
  \end{tabular}
\end{table}

\providecommand{\cutoff}{cutoff}
\providecommand{\Cutoff}{Cutoff}
\providecommand{\cutoffs}{cutoffs}
\providecommand{\Cutoffs}{Cutoffs}

\vspace{-0.9\baselineskip}
\subsection{Router Results: \Cutoff{} vs.\ Baseline}
\label{subsec:router-results}

\textbf{Setup.}
We compare the router’s \emph{final decision stage} with and without a \cutoff{} on the tuning splits
(\textbf{training}: 102, \textbf{held-out}: 18).
For each image, the router emits a \emph{predicted label} (either a \texttt{MODEL\_\#} or \texttt{None}) together with a
\emph{predicted confidence} in $[0,1]$, i.e., the probability the router assigns to the emitted label at decision time.
Accuracy is measured by exact match between the \emph{predicted label} and the \emph{ground-truth label}.
Calibration is summarized with the \textbf{Expected Calibration Error (ECE)} using 10 equal-width, right-closed bins:
\[
\mathrm{ECE} \;=\; \sum_{b=1}^{10} \frac{n_b}{N}\,\bigl|\mathrm{acc}_b-\mathrm{conf}_b\bigr| \quad\text{(lower is better).}
\]


\vspace{-0.5\baselineskip} 
\begin{table}[!ht]
  \centering
  \scriptsize
  \setlength{\tabcolsep}{2pt}
  \caption{Final decision stage results. Accuracy is exact match;
    ECE uses 10 equal-width, right-closed bins.}
  \label{tab:router-metrics}
  \begin{tabular}{@{}P{0.55\linewidth}cc@{}}
    \toprule
    \textbf{Split / Method} & \textbf{Accuracy} & \textbf{ECE} \\
    \midrule
    Training — Baseline (no \cutoff) & 0.392 & 0.415 \\
    Training — \Cutoff{} (ours)      & \textbf{0.480} & 0.490 \\
    \midrule
    Held-out — Baseline (no \cutoff) & 0.444 & 0.390 \\
    Held-out — \Cutoff{} (ours)      & \textbf{0.556} & \textbf{0.372} \\
    \bottomrule
  \end{tabular}
\end{table}
\vspace{0.1\baselineskip} 

\begin{figure}[t]
  \centering
  \includegraphics[width=\columnwidth]{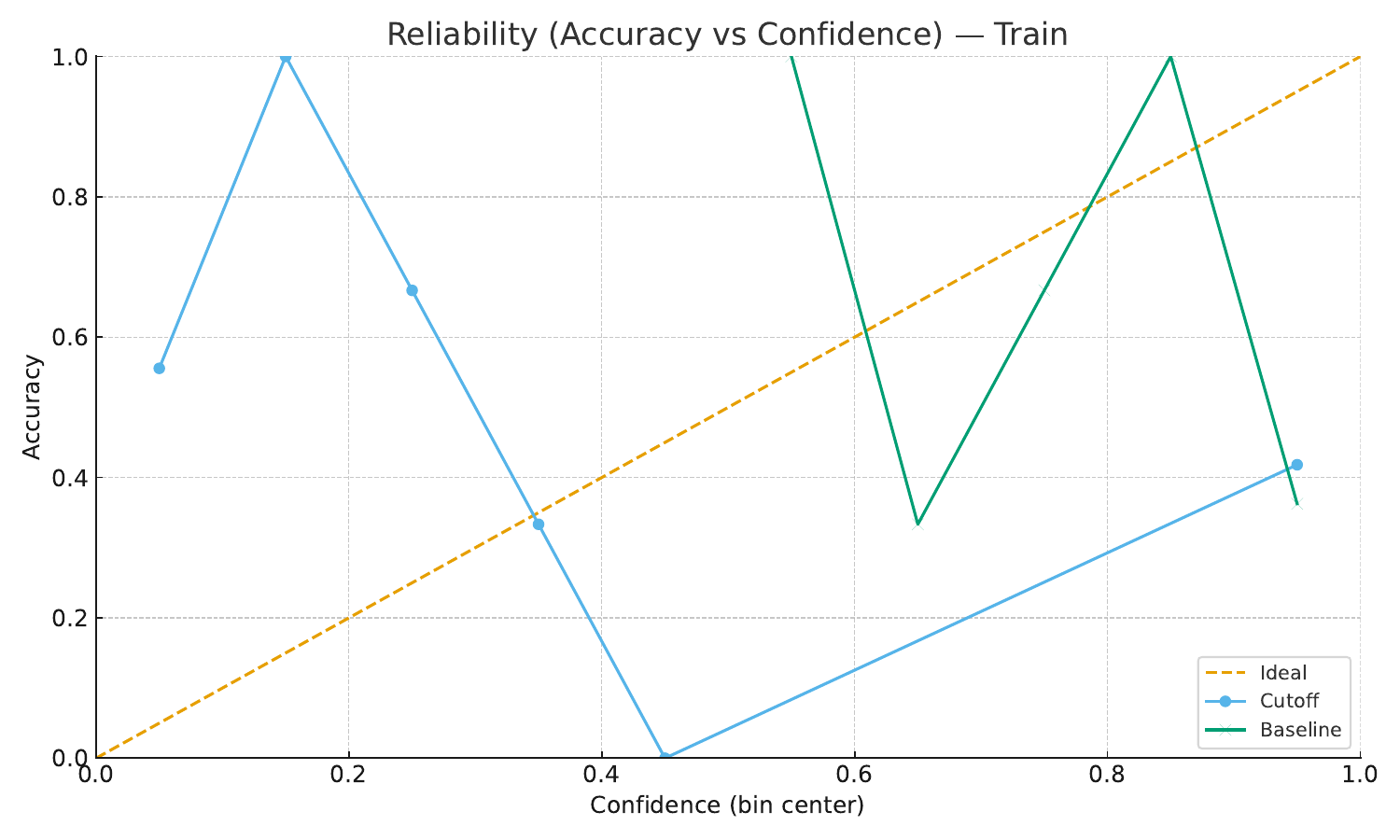}
  \caption{Reliability (Accuracy vs.\ Confidence) on \textbf{training split}. Dashed line: ideal $y{=}x$.
  The \cutoff{} router improves accuracy but is slightly under-calibrated in mid/high-confidence bins.}
  \label{fig:rel-train}
\end{figure}

\begin{figure}[t]
  \centering
  \includegraphics[width=\columnwidth]{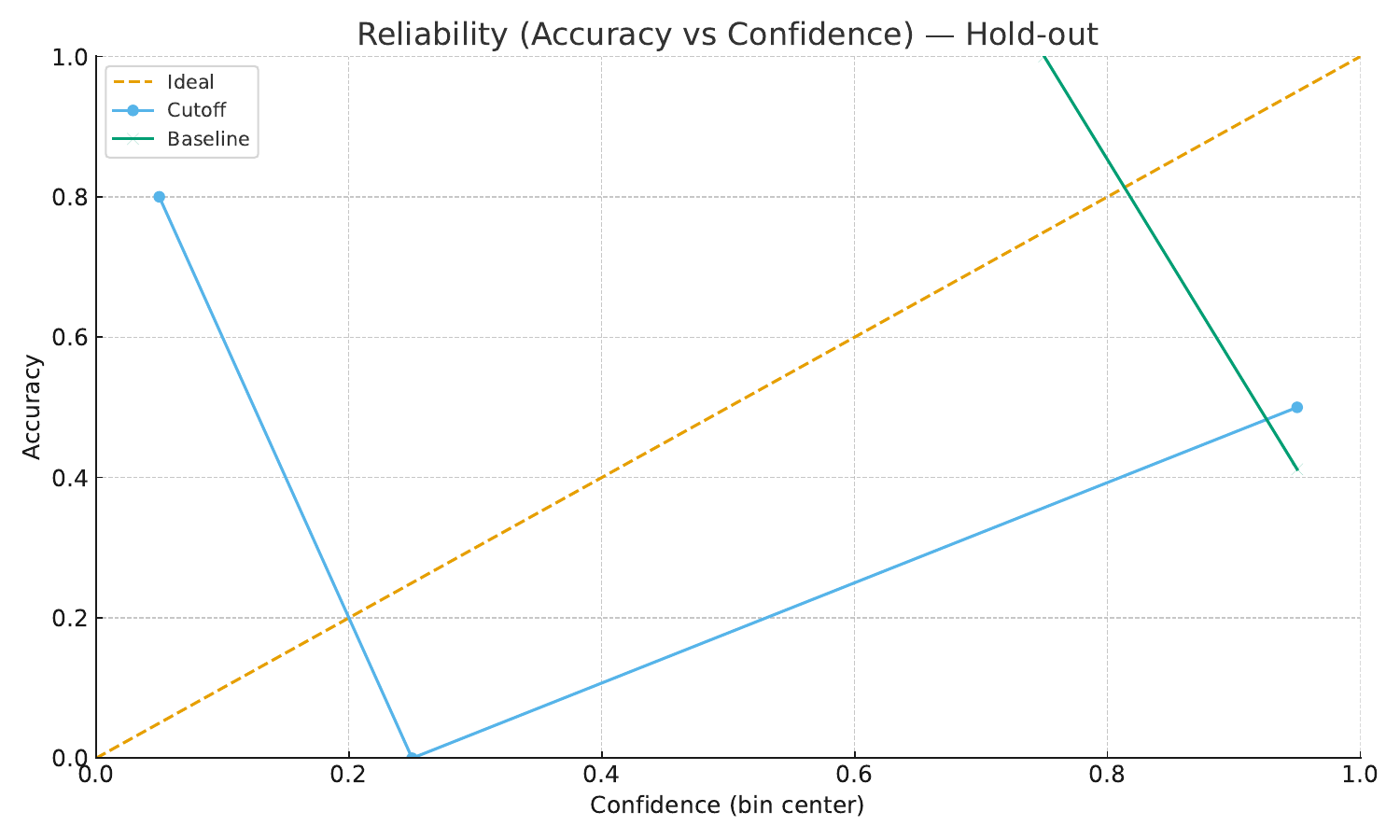}
  \caption{Reliability (Accuracy vs.\ Confidence) on \textbf{held-out}. The \cutoff{} router improves both accuracy and calibration
  (lower ECE) compared to the baseline.}
  \label{fig:rel-hold}
\end{figure}

\textbf{Reliability curves.}
Figures~\ref{fig:rel-train} and~\ref{fig:rel-hold} plot \emph{accuracy vs.\ confidence} per bin for the \cutoff{} router and the
baseline, along with the ideal $y{=}x$ diagonal. On the held-out split, the \cutoff{} curve lies closer to the diagonal (lower ECE).
On the training split, the \cutoff{} router improves accuracy but is mildly under-calibrated at mid/high confidence, increasing ECE. Overall, \Cutoff{} improves end-task correctness on both splits (+8.8 pp on training split; +11.1 pp on held-out split) and achieves better held-out calibration (ECE $0.372$ vs.\ $0.390$). A practical advantage, visible in Figures~\ref{fig:rel-train} and~\ref{fig:rel-hold}, is that when confidence exceeds $0.5$, accuracy under \Cutoff{} tends to increase monotonically with confidence; in contrast, the baseline exhibits mixed behavior, where confidence does not reliably track accuracy. Furthermore, on the held-out split the \cutoff{} curve closely mirrors the shape observed on the training split—aside from a slight deviation in the first (lowest-confidence) range—indicating that the reliability profile measured on the training data largely transfers to unseen data. In practice, this suggests that development-time reliability estimates are predictive of deployment-time reliability.

\section{Solution 2: Specialty-Specific Model for Downstream Deployment}
Building on the rationale in Section~\ref{intro} (Solution~2), we fine-tune the same backbone VLM adopted in Section~\ref{sol1}—\textsc{MedGemma}—across five specialties (gastroenterology, hematology, ophthalmology, pathology, and radiology) to simplify maintenance and deployment rather than pursuing state-of-the-art performance on each task. Importantly, Solution~2 is implemented and evaluated independently; it does not consume outputs from Solution~1, though the two are conceptually complementary.

For each specialty we compare: (a) an external task-specific benchmark (typically the published state-of-the-art for that dataset), (b) the base MedGemma in zero-shot mode (no domain adaptation), and (c) MedGemma after specialty-specific fine-tuning (denoted “S–S” in the table). This tests whether a single specialty-tuned MedGemma can substitute for the external benchmark while retaining a single deployment stack. If successful, a fixed VLM plus targeted prompt engineering and supervised fine-tuning may reduce the need to design, validate, and operate separate architectures—saving compute and shortening triage-to-deployment cycles~\citet{annarumma2019triage}. All results are summarized in Table~\ref{tab:overall_summary_final}.

\paragraph{\textbf{Datasets}}
\label{dataset}
For gastroenterology, we use the dataset introduced by~\citet{li2021colonoscopy}, an annotated compilation of three public sources: CVC-ColonDB~\citet{bernal2012towards}, GLRC~\citet{mesejo2016computer}, and KUMC (80 colonoscopy video sequences from the University of Kansas Medical Center). For hematology, we use BloodMNIST~\citet{acevedo2020dataset}, organized into eight blood cell categories. For ophthalmology, we employ OCTMNIST~\citet{kermany2018identifying} (four retinal OCT classes) and RetinaMNIST~\citet{dataset20202nd} (five retinopathy grades). For pathology, we use TissueMNIST~\citet{ljosa2012annotated} (eight kidney cortex cell types), PathMNIST~\citet{kather2019predicting} (nine colorectal tissue types), and a breast cancer histopathology set following~\citet{cruz2014automatic}. Lastly, for radiology, we use BreastMNIST~\citet{al2020dataset} (binary classification of breast images: normal vs.\ malignant), DeepLesion~\citet{yan2017deeplesion} (eight different body organs), Organ(A-B-C)MNIST~\citet{bilic2023liver, xu2019efficient} (11 body organs in axial, coronal, and sagittal views, respectively), PneumoniaMNIST~\citet{kermany2018identifying, kermany2018large} (pneumonia vs.\ normal chest images), and ChestMNIST~\cite{wang2017chestx} (14 disease labels in X-ray images).
Train/validation/test sizes are summarized in Table~\ref{tab:overall_summary_final}.

\paragraph{\textbf{Fine-tuning results}}
For each \emph{specialty-specific} case (five in total), we use a parameter-efficient fine-tuning (PEFT) method, QLoRA~\citet{dettmers2023qlora}. As shown in Table~\ref{tab:overall_summary_final}, specialty-specific fine-tuning substantially narrows the gap to specialist benchmarks. While the base MedGemma performs modestly in zero-shot mode, targeted specialty adaptation yields competitive results across modalities and objectives. This supports the premise that one VLM can act as a generalist engine per specialty, reducing the need for separate per-task architectures and thereby saving compute and shortening triage-to-deployment cycles. We also provide public example inference results for each specialty-specific fine-tuning case \href{https://osf.io/2jebh/overview?view_only=9caffb08ae4444e0952916ba232cb8f3}{here}.

\section{Discussion}
We presented two linked clinical solutions: (1) a single, calibrated medical VLM that runs an aware model-card–matching workflow, routing an input image to a model-card ID—or abstaining and terminating when appropriate—and (2) the same VLM used as a specialty-specific downstream model with performance competitive with task-specific models. The solutions are complementary but \emph{decoupled} in our implementation: Solution~2 neither requires nor consumes Solution~1 outputs.
For \textbf{Solution~1}, the workflow is designed to streamline selection for data scientists by providing: \emph{safety and trust} (stage-wise decisions are visible throughout the pipeline), \emph{robustness} (fewer incorrect selections via the combined use of stage-wise prompts and answer-selector logic), \emph{scalability} (flexibility to add stages, descriptors, or model cards, or to swap in a more mature selector), and \emph{speed} (a minimalist design with fewer moving parts than recent multi-agent pipelines~\citep{shimgekar2025agentic,202508.1713}).
For \textbf{Solution~2}, specialty-specific models offer a practical first step toward automated deployment. We make this step concrete by shifting effort from designing new architectures to engineering task-specific prompts that align the VLM’s outputs with downstream objectives. In practice, this process can increase throughput for data scientists and clinicians without sacrificing performance.

\textbf{Limitations and future work.}
MedGemma is pretrained on a fixed set of modalities and tasks; accordingly, it may not generalize to imaging types or findings outside its training distribution. In short, its inference and selection decisions are tied to its training data: if a modality or pathology is unseen or underrepresented, the model may fail to recognize it and either choose an imperfect model or default to ``Other/None.'' This risk can be mitigated by enlarging the pretraining corpus or by adopting continual-learning methods that add new domains while preserving prior knowledge (e.g., \citet{li2017learning}).

While our system outputs justifications for model selection, the interpretability of the VLM’s internal reasoning remains limited (a common issue with large neural models). Techniques such as chain-of-thought prompting or explicability constraints could be applied so the model provides brief, stage-wise reasoning—further improving transparency (see \citet{wei2022cot}). Another design choice concerns the selection of thresholds \( \tau_{1}, \tau_{2}, \tau_{3} \). We derive the optimal cutoff thresholds using Algorithm~\ref{alg:cutoff-sweep}, which maximizes the per-stage score given (i) our stage-specific prompting strategy; (ii) the diversity and structure of the model-card repository defined in Section~\ref{sol1}; and (iii) the workflow’s preference for the true final output—either \texttt{None} or a model-card ID—across different input images. In practice, these thresholds may require adjustment across clinical contexts. For high-accuracy applications using the same modality and disease labels as the training set, a workflow designer might prefer \texttt{None} over the closest model-card match, resulting in different labels for Stage~3 tuning.

Finally, our pipeline has three stages because repository lookup is primarily based on modality and a single primary finding. As we incorporate more complex model cards (e.g., some models might handle combinations of findings or require patient metadata), the prompt design may need to extend to additional stages or a more free-form query. Fortunately, the answer-selector concept generalizes to any number of stages by examining the confidence of alternative responses. We outline two complementary directions for deployment: (1) automate fine-tuning end to end—so deployment proceeds with minimal data-scientist intervention—by automatically selecting the dataset linked to the model card chosen in Solution~1 and automatically generating the prompts required for fine-tuning; and (2) reduce inference latency by applying pruning methods tailored to large language models (e.g., SparseGPT~\citet{frantar2023sparsegpt}).

\clearpage
\begin{table*}[p]
\caption{Performance summary across specialties}
\label{tab:overall_summary_final}
\centering
\begingroup
\setlength{\tabcolsep}{25pt}
\renewcommand{\arraystretch}{0.95}
\footnotesize

\scalebox{\tablescale}{%
\begin{tabular}{@{}%
  L{0.450\textwidth}  
  C{0.135\textwidth}  
  C{0.110\textwidth}  
  C{0.160\textwidth}  
@{}}
\toprule
\textbf{Task} &
\thead{External\\benchmark} &
\thead{MedGemma} &
\thead{(S\textendash S)\\fine-tuned} \\
\midrule

\multicolumn{4}{@{}l@{}}{\textbf{Gastroenterology} — \emph{PolypSet} {\footnotesize(Training: 27{,}048; Test: 4{,}719; Val: 4{,}214)}}\\[-1pt]
\cmidrule(l{2pt}r{2pt}){1-4}
Polyp Classification (Acc.)            & \textbf{81}\%  & 61\% & 79\%  \\
Polyp Classification (F1 Macro)        & \textbf{80}\%  & 38\% & 77\%  \\
Polyp Classification (F1 Weighted)     & \textbf{81}\%  & 47\% & 79\%  \\
BBox Localization Acc.                 & \textbf{78}\%  &  1\% & \NA   \\
Region Localization Acc.               & \NA   & 32\% & \textbf{46}\%  \\
\addlinespace[2pt]

\cmidrule(l{2pt}r{2pt}){1-4}
\multicolumn{4}{@{}l@{}}{\textbf{Hematology} — \emph{BloodMNIST} {\footnotesize(Training: 11{,}959; Test: 3{,}421; Val: 1{,}712)}}\\[-1pt]
\cmidrule(l{2pt}r{2pt}){1-4}
Cell-type Classification (Acc.)        & 97\%  & 20\% & \textbf{98}\%  \\
Cell-type Classification (F1 Macro)    & \NA   &  8\% & \textbf{98}\%  \\
Cell-type Classification (F1 Weighted) & \NA   & 10\% & \textbf{98}\%  \\
\addlinespace[2pt]

\cmidrule(l{2pt}r{2pt}){1-4}
\multicolumn{4}{@{}l@{}}{\textbf{Ophthalmology} — \emph{OCTMNIST} {\footnotesize(Training: 97{,}477; Test: 1{,}000; Val: 10{,}832)}}\\[-1pt]
\cmidrule(l{2pt}r{2pt}){1-4}
Retinal OCT Classification (Acc.)      & 78\%  & 26\% & \textbf{93}\% \\
Retinal OCT Classification (F1 Macro)  & \NA   & 38\% & \textbf{77}\% \\
Retinal OCT Classification (F1 Weighted)& \NA  & 47\% & \textbf{79}\% \\
\addlinespace[1pt]

\cmidrule(l{2pt}r{2pt}){1-4}
\multicolumn{4}{@{}l@{}}{\textbf{Ophthalmology} — \emph{RetinaMNIST} {\footnotesize(Training: 1{,}080; Test: 400; Val: 120)}}\\[-1pt]
\cmidrule(l{2pt}r{2pt}){1-4}
Diabetic Retinopathy Grading (Acc.)    & 53\%  & 55\% & \textbf{74}\% \\
Diabetic Retinopathy Grading (F1 Macro)& \NA   & 37\% & \textbf{59}\% \\
Diabetic Retinopathy Grading (F1 Weighted)& \NA & 52\% & \textbf{72}\% \\
\addlinespace[2pt]

\cmidrule(l{2pt}r{2pt}){1-4}
\multicolumn{4}{@{}l@{}}{\textbf{Pathology} — \emph{TissueMNIST} {\footnotesize(Training: 165{,}466; Test: 47{,}280; Val: 23{,}640)}}\\[-1pt]
\cmidrule(l{2pt}r{2pt}){1-4}
Kidney Cortex Classification (Acc.)    & \textbf{70.3}\%& 21\% & 70.1\% \\
Kidney Cortex Classification (F1 Macro)& \NA   &  5\% & \textbf{60}\% \\
Kidney Cortex Classification (F1 Weighted)& \NA&  9\% & \textbf{69}\% \\
\addlinespace[1pt]

\cmidrule(l{2pt}r{2pt}){1-4}
\multicolumn{4}{@{}l@{}}{\textbf{Pathology} — \emph{PathMNIST} {\footnotesize(Training: 89{,}996; Test: 7{,}180; Val: 10{,}004)}}\\[-1pt]
\cmidrule(l{2pt}r{2pt}){1-4}
Tissue Type Classification (Acc.)      & 91.1\%& 38\% & \textbf{96}\% \\
Tissue Type Classification (F1 Macro)  & \NA   & 28\% & \textbf{94}\% \\
Tissue Type Classification (F1 Weighted)& \NA  & 29\% & \textbf{96}\% \\
\addlinespace[1pt]

\cmidrule(l{2pt}r{2pt}){1-4}
\multicolumn{4}{@{}l@{}}{\textbf{Pathology} — \emph{BreastHIST} {\footnotesize(Training: 214{,}969; Test: 31{,}376; Val: 31{,}179)}}\\[-1pt]
\cmidrule(l{2pt}r{2pt}){1-4}
Ductal Carcinoma Classification (Acc.) & 84.3\%& 69\% & \textbf{87}\% \\
Ductal Carcinoma Classification (F1 Macro)& \NA& 66\% & \textbf{84}\% \\
Ductal Carcinoma Classification (F1 Weighted)& \NA& 70\% & \textbf{87}\% \\
\addlinespace[1pt]

\cmidrule(l{2pt}r{2pt}){1-4}
\multicolumn{4}{@{}l@{}}{\textbf{Radiology} — \emph{BreastMNIST} {\footnotesize(Training: 546; Test: 156; Val: 78)}}\\[-1pt]
\cmidrule(l{2pt}r{2pt}){1-4}
Breast Lesion Classification (Acc.) & \textbf{86.3}\%& 73\% & 74\% \\
Breast Lesion Classification (F1 Macro)& \NA& 42\% & \textbf{52}\% \\
Breast Lesion Classification (F1 Weighted)& \NA& 62\% & \textbf{67}\% \\
\addlinespace[1pt]
\cmidrule(l{2pt}r{2pt}){1-4}
\multicolumn{4}{@{}l@{}}{\textbf{Radiology} — \emph{DeepLesion} {\footnotesize(Training: 7{,}859; Test: 984; Val: 973)}}\\[-1pt]
\cmidrule(l{2pt}r{2pt}){1-4}
BBox-Guided Region Classification (Acc.) & \NA& 28\% & \textbf{90}\% \\
BBox-Guided Region Classification (F1 Macro)& \NA& 19\% & \textbf{87}\% \\
BBox Guided Region Classification (F1 Weighted)& \NA& 23\% & \textbf{90}\% \\
\addlinespace[1pt]
\cmidrule(l{2pt}r{2pt}){1-4}
\multicolumn{4}{@{}l@{}}{\textbf{Radiology} — \emph{OrganAMNIST} {\footnotesize(Training: 34{,}859; Test: 17{,}778; Val: 6{,}491)}}\\[-1pt]
\cmidrule(l{2pt}r{2pt}){1-4}
Organ Classification (Axial) (Acc.) & \textbf{95}\% & 9\% & 94\% \\
Organ Classification (Axial) (F1 Macro)& \NA& 4\% & \textbf{92}\% \\
Organ Classification (Axial) (F1 Weighted)& \NA& 6\% & \textbf{94}\% \\
\addlinespace[1pt]
\cmidrule(l{2pt}r{2pt}){1-4}
\multicolumn{4}{@{}l@{}}{\textbf{Radiology} — \emph{OrganCMNIST} {\footnotesize(Training: 12{,}975; Test: 8{,}216; Val: 2{,}392)}}\\[-1pt]
\cmidrule(l{2pt}r{2pt}){1-4}
Organ Classification (Coronal) (Acc.) & \textbf{92}\% & 10\% & 88\% \\
Organ Classification (Coronal) (F1 Macro)& \NA& 7\% & \textbf{86}\% \\
Organ Classification (Coronal) (F1 Weighted)& \NA& 9\% & \textbf{88}\% \\
\addlinespace[1pt]
\cmidrule(l{2pt}r{2pt}){1-4}
\multicolumn{4}{@{}l@{}}{\textbf{Radiology} — \emph{OrganSMNIST} {\footnotesize(Training: 13{,}932; Test: 8{,}827; Val: 2{,}452)}}\\[-1pt]
\cmidrule(l{2pt}r{2pt}){1-4}
Organ Classification (sagittal) (Acc.) & \textbf{81}\% & 7\% & 75\% \\
Organ Classification (sagittal) (F1 Macro)& \NA& 3\% & \textbf{93}\% \\
Organ Classification (sagittal) (F1 Weighted)& \NA& 3\% & \textbf{95}\% \\
\addlinespace[1pt]
\cmidrule(l{2pt}r{2pt}){1-4}
\multicolumn{4}{@{}l@{}}{\textbf{Radiology} — \emph{PneumoniaMNIST} {\footnotesize(Training: 4{,}708; Test: 624; Val: 524)}}\\[-1pt]
\cmidrule(l{2pt}r{2pt}){1-4}
Chest Pneumonia Classification (Acc.) & 95\% & 28\% & \textbf{95}\% \\
Chest Pneumonia Classification (F1 Macro)& \NA& 73\% & \textbf{94}\% \\
Chest Pneumonia Classification (F1 Weighted)& \NA& 74\% & \textbf{94}\% \\
\addlinespace[1pt]
\cmidrule(l{2pt}r{2pt}){1-4}
\multicolumn{4}{@{}l@{}}{\textbf{Radiology} — \emph{ChestMNIST} {\footnotesize(Training: 78{,}468; Test: 22{,}433; Val: 11{,}219)}}\\[-1pt]
\cmidrule(l{2pt}r{2pt}){1-4}
Chest Multi-label Classification (Acc.) & \textbf{95}\% & 86\% & 92\% \\
Chest Multi-label Classification (F1 Macro)& \NA& \textbf{16}\% & 14\% \\
Chest Multi-label Classification (F1 Weighted)& \NA& 40\% & \textbf{40}\%
\\
\bottomrule
\end{tabular}%
} 
\endgroup
\end{table*}
\clearpage 

\bibliography{jmlr-sample}

\end{document}